\documentclass{article}
\pdfoutput=1 

\usepackage{natbib}
\setcitestyle{numbers,square}

\usepackage{graphicx}
\usepackage{subfig}
\usepackage{bbding} 
\usepackage{makecell}
\usepackage{amsmath}
\usepackage{wrapfig}
\usepackage{picinpar}
\usepackage{cutwin}
\usepackage{multirow}
\usepackage{tabularx}
\usepackage{diagbox}
\usepackage{colortbl}
\usepackage[preprint]{neurips_2024}

\usepackage[utf8]{inputenc} 
\usepackage[T1]{fontenc}    
\usepackage{hyperref}       
\usepackage{url}            
\usepackage{booktabs}       
\usepackage{amsfonts}       
\usepackage{nicefrac}       
\usepackage{microtype}      
\usepackage{xcolor}         

\title{Debiased Prompt Tuning in Vision-Language Model without Annotations}

\author{
  Chaoquan Jiang$^{1}$, Yunfan Yang$^{1}$, Rui Hu$^{1}$, Jitao Sang$^{1}$ \\
  $^{1}$School of  Computer  and  Information Technology \\
  \& Beijing Key Lab  of Traffic Data Analysis and  Mining\\
  Beijing  Jiaotong University,  China\\
  \texttt{\{cqjiang,yunfanyang,ruihu,jtsang\}@bjtu.edu.cn}\\
}

\begin{document}

\maketitle

\begin{abstract}
Prompt tuning of Vision-Language Models (VLMs) such as CLIP, has demonstrated the ability to rapidly adapt to various downstream tasks. However, recent studies indicate that tuned VLMs may suffer from the problem of spurious correlations, where the model relies on spurious features (e.g. background and gender) in the data. This may lead to the model having worse robustness in out-of-distribution data. Standard methods for eliminating spurious correlation typically require us to know the spurious attribute labels of each sample, which is hard in the real world. In this work, we explore improving the group robustness of prompt tuning in VLMs without relying on manual annotation of spurious features. We notice the zero - shot image recognition ability of VLMs and use this ability to identify spurious features, thus avoiding the cost of manual annotation. By leveraging pseudo-spurious attribute annotations, we further propose a method to automatically adjust the training weights of different groups. Extensive experiments show that our approach efficiently improves the worst-group accuracy on CelebA, Waterbirds, and MetaShift datasets, achieving the best robustness gap between the worst-group accuracy and the overall accuracy. 
\end{abstract}

\section{Introduction}
\label{sec:intro}
Recently, pre-trained Vision-Language Models (VLMs), such as CLIP\cite{radford2021learning} and ALIGN\cite{jia2021scaling}, have demonstrated impressive capabilities in various downstream tasks, including image classification\cite{zhou2022learning,singh2022flava}, visual understanding\cite{gu2021open,li2021align} and object detection\cite{du2022learning}, etc. Benefiting from rich representations through pre-training on large-scale image-text data, prompt tuning\cite{zhou2022learning,zhou2022conditional} has been the dominant methodology to improving the performance of VLMs on downstream tasks. Among these attempts, CoOp\cite{zhou2022learning} trains learnable vectors that contain class names to adjust textual representations and greatly improves the performance. Prompt tuning avoids fine-tuning the entire model and thus enables the model to adapt efficiently to downstream tasks\cite{zhou2022learning,zhu2023debiased,yao2023visual}.

However, we have noticed that prompt-tuned VLMs also face the problem of relying on spurious correlations. That is, the model depends on features in the data that are spuriously related to the labels, such as background and gender\cite{geirhos2020shortcut}, resulting in poor performance on minority groups within the data. For example, in the Waterbirds dataset\cite{sagawa2019distributionally}, where the task is to distinguish between land birds and water birds, due to the spurious correlation between the target and the land or ocean background, the accuracy of "water birds on land" and "land birds on the water" significantly declines after prompt tuning\cite{zhangamend}. Moreover, in real-world applications, relying on spurious correlations may cause serious safety issues and societal impacts, such as medical diagnosis\cite{oakden2020hidden,kermany2018identifying}, autonomous driving\cite{cui2019multimodal}, and face  recognition\cite{seo2022unsupervised}. 

Many works focus on eliminating the reliance on spurious correlations when training visual models, while lacking exploration of prompt tuning for VLMs. Previous methods require obtaining the class labels and spurious attribute labels of the training data \cite{sagawa2019distributionally,idrissi2022simple,lu2021invariant,zhu2021learning}. However, this potentially  requires manual annotation of the spurious attributes of each sample, which is costly for large datasets. In addition, these conventional methods require fine-tuning the entire visual model, it is computationally expensive to apply them to the foundation VLMs. The latest works focus on fine-tuning a small part of parameters of VLMs, but they are either computationally complex \cite{zhang2022contrastive} or lack effectiveness \cite{you2024calibrating,chuang2023debiasing,zhangamend}. Currently, eliminating the spurious correlations of VLMs in prompt tuning remains underexplored. 

This work focuses on improving robustness of fine-tuning VLMs without relying on manually annotated spurious attribute. Our work proposes two objectives: \textbf{(1) Automatically annotate specific spurious attributes.}  Recent research has explored reducing the reliance on spurious attribute annotation. For example, clustering deep representations \cite{seo2022unsupervised} or leveraging the fine-tuned model's predictions \cite{nam2020learning,liu2021just}  to generate pseudo-labels. However, due to the lack of explicit supervision, these methods do not always effectively annotate specific attributes. Inspired by the zero-shot capabilities of VLMs \cite{radford2021learning}, we propose using VLMs to infer specific spurious attributes through simple language descriptions. \textbf{(2) Robust prompt-tuning for VLMs.} To prevent the prompt tuning from obtaining a biased textual representations, we proposes a re-weighting method \textbf{DPT}, a \textbf{D}ebiased \textbf{P}rompt-\textbf{T}uning method that can  automatically adjust the training weights of different groups. Extensive experiments show that DPT achieves best performance on three datasets: Waterbirds, CelebA, and MetaShift. In particular, compared with methods supervised by group annotations, we achieve competitive results.

Our contributions are summarized as follows:
\begin{itemize}
    \item We propose using the zero-shot capabilities of VLMs to directly infer predefined spurious attribute labels, which is more effective than the unsupervised methods ;
    \item We present a robust prompt-tuning method, demonstrating that the pre-trained visual representations of VLMs are sufficient to train a robust classifier, achieving good performance on multiple tasks;
    \item Our method can also be used to improve the robustness of unimodal visual models, thus having broader practical value.
\end{itemize}

\section{Related Works}
\label{sec:related_work}
\paragraph{Mitigating Spurious Correlations}
There is a growing number of works focusing on eliminating spurious correlations, but many of them only address single-modality vision models. In computer vision, models typically rely on the irrelevant attributes such as an image's background \cite{moayeri2022comprehensive,xiao2020noise}, gender and racial \cite{karkkainen2021fairface,park2022fair}, and other features \cite{geirhos2020shortcut,geirhos2018imagenet}. As the models are applied to high-risk fields such as medical images \cite{oakden2020hidden}, autonomous driving \cite{cui2019multimodal}, and face recognition \cite{park2022fair}, concerns about the model's robustness and fairness have been raised. To mitigate spurious correlation, many studies have proposed solutions: reweighting or resampling \cite{sagawa2019distributionally,idrissi2022simple}, adversarial training \cite{lu2021invariant,zhu2021learning}, contrastive learning\cite{park2022fair,wang2022counterexample}, feature disentangling\cite{lee2021learning},etc. However, these methods require fine-tuning the entire visual model, even more than once, and cannot be effectively applied to the prompt tuning of frozen image and text encoders. In addition, these methods potentially require annotation of spurious attributes, which can be costly and difficult to apply on large datasets.

\paragraph{Robustness without Group Annotations}
Some works have been devoted to eliminating the dependence on annotating spurious attributes. SSA\cite{nam2022spread} and DFR\cite{kirichenko2022last} propose a semi-supervised scenario, requiring that some subsets of the training set have annotation of spurious attributes. SSA is used to label the model with semi-supervised training data, and DFR is used to obtain a balanced data subset to train a robust classifier. A further setting is that all samples  have no any spurious attribute labeling in the training set. The main solution is a two-stage training: in the first stage, the model is fitted with empirical risk minimization to obtain difficult\cite{nam2020learning,qiu2023simple,zhang2022correct} or misclassified examples\cite{liu2021just,nam2020learning}; in the second stage, these samples are reweighted to improve robustness. However, the first stage requires fine-tuning the entire model, which may lead to overfitting or underfitting the training set and may be computationally expensive when applied to large-scale foundation VLMs. In addition, some works propose using pre-trained representations for unsupervised clustering\cite{sohoni2020no,seo2022unsupervised} to obtain group labels. However, the above unsupervised labeling methods cannot accurately label specific spurious attributes.

\paragraph{Improve Group Robustness of VLMs}
The group robustness of Vision-Language Models (VLMs) has received increasing attention. \cite{zhang2022contrastive} investigated the zero-shot ability of VLMs and discussed poor group robustness on many tasks; \cite{agarwal2021evaluating,berg2022prompt} studies also showed that VLMs have bias. However, due to the large number of parameters in the foundation VLMs, the mainstream work focuses on training or fine-tuning a small number of parameters to alleviate the bias problem of VLMs\cite{berg2022prompt,zhang2022contrastive,you2024calibrating,yang2023mitigating}. \cite{zhang2022contrastive} improved the group robustness by training the adapter with contrastive learning based on the pre-trained representation; \cite{you2024calibrating} fine-tuned the last attention layer of the visual encoder to calibrate the visual representations; \cite{seth2023dear} proposed to eliminate the spurious correlation through learn additive residual image representations. These methods require adjusting the visual representations, while our work shows that the pre-trained representation is sufficient to train a robust classifier. In addition, some work introduced text representations to mitigate spurious correlation of VLMs. \cite{berg2022prompt} proposed to perform prompt learning in an adversarial way; \cite{an2023more} proposed to first infer the task-irrelevant context to improve the robustness; \cite{chuang2023debiasing} proposed to project the biased direction in the text space. Directly, we utilized the text representations of VLMs to first infer the spurious attribute labels and only train the text prompt to improve the group robustness.

\section{Methods}
\label{sec:method}
\begin{figure*}[htbp]
\centering
\includegraphics[width=1.0\linewidth]{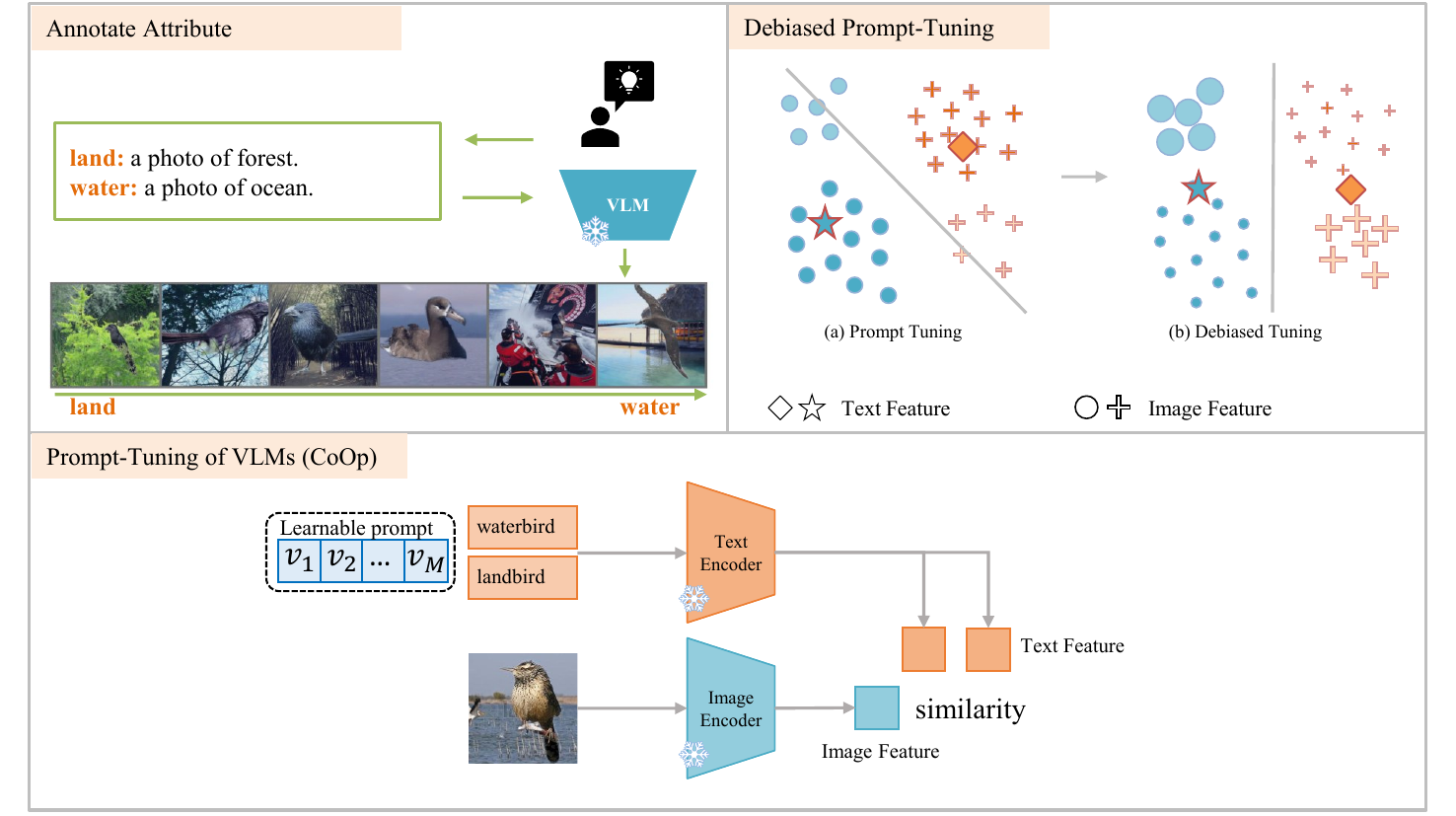}
  \caption{The framework of Debiased Prompt-Tuning. On the top-left, we first use VLMs and text descriptions to infer spurious attribute labels and get pseudo-group annotations. On the top-right, misclassified groups are automatically re-weighted, and unbiased text features are obtained by prompt tuning. On the bottom, we revisit the prompt tuning method (Context Optimization) of VLMs and get the image and text features. 
  }
  \label{fig:head_shot}
\end{figure*}
\subsection{Preliminaries}
\label{sec:preliminary}
\paragraph{Setting} We consider the setting of group robustness, following the mainstream practice \cite{sohoni2020no,sagawa2019distributionally,seo2022unsupervised,kirichenko2022last}. We denote the input image is \(x\in X\) and the label is \(y\in Y\). In particular, we assume that the data distribution consists of several groups \(g\in G\), and these groups are combinations of the target label \(y\) and the spurious attribute \(s\in G\), symbolically represented as \(g\in Y\times S\). Take the waterbird dataset\cite{sagawa2019distributionally} as an example, we classify birds into \(\{landbird, waterbird\}\), and the spurious attribute is the background of the picture \(\{land, water\}\), which combines into four groups. In the dataset, there is spurious correlations between the attribute \(s\) and the label \(y\), about 95\% of the image backgrounds of landbirds appear on land, and about 95\% of waterbirds appear on water. As a result, the classifier trained on such spuriously correlated data distribution may rely heavily on the background and perform poorly on the two subgroups of \(\{waterbird\ on\ land\}\) and \(\{landbird\ on\ water\}\).

To evaluate the model's robustness to spurious correlations, we follow the classic evaluation setting\cite{sagawa2019distributionally,zhang2022contrastive}, and use two metrics, Worst-group accuracy (WG) and the accuracy gap (GAP) between worst-group accuracy and overall accuracy.

\paragraph{CLIP} Contrastive Vision-Language Models\cite{radford2021learning} conduct contrastive learning on 400M image-text data, encoding visual and text representations into an aligned space. CLIP consists of a visual encoder \(f_v\) and a text encoder \(f_t\). Suppose there is an input image \(x\) and \(K\) text prompts for categories, such as "a photo of a [class]". With the prompts into the frozen text encoder, we can obtain text embedding vectors \(\{w_j\}_{j = 1}^K\). Zero-shot classification of the image can be completed by calculating the cosine similarity between the visual embedding vector \(f_v(x)\) and the text embedding vectors \(\{w_j\}_{j = 1}^K\). The classification probability is as follows:
\begin{equation}
p_{\theta}(y=j|x)=\frac{exp(cos(f_v(x),w_j)/\tau)}{\sum_y exp(cos(f_v(x),w_y)/\tau)},
\end{equation}
here, \(\tau\) is the temperature parameter, which is set to 30 by default in this paper; \(\cos(\cdot ,\cdot)\) represents the cosine similarity. 

\paragraph{Context Optimization (CoOp)} Prompt learning\cite{yao2023visual} is a popular parameter-efficient fine-tuning method for adapting vision-language models to downstream tasks. The classic CoOp trains learnable text vectors and the image/text encoder remains frozen during the prompt learning process. A Soft prompt can be defined as follows:
\begin{equation}
    t_i=[v_1, v_2,\cdots, v_M, c_i],
\end{equation}
here, \(i\in\{1,\cdots,K\}\) is the class index, \(c_i\) is the word embedding of the \(i_{th}\) class, and \(v_j\) is a learnable vector with the same dimension (i.e., 512 for CLIP) as the word embedding, which is randomly initialized. Similar to manual templates of zero-shot classification, the text embedding \(w_i = f_t(t_i)\) of each class's soft prompt can be obtained. Then, maximizing the probability of the sample and the text embedding of the target category can improve the performance in downstream tasks. 

In this work, the CoOp method is applied to the training of downstream tasks for improving group robustness.  In particular, on the benchmark of group robustness, we propose to pre-extract the visual features of all training samples, and use the above-mentioned prediction probabilities and the robust re-weighted loss we proposed to train the soft prompt, so as to obtain a robust classifier.
\subsection{Overview of DPT}
In this section, we introduce Debiased Prompt Tuning (DPT), as shown in Fig. \ref{fig:head_shot}, a prompt tuning method, aimed at improving group robustness of CLIP models, relying on VLMs' zero-shot recognition ability to directly infer spurious attribute labels. Further, in order to efficiently improve the group robustness of VLMs, we propose a strategy for automatically weighting samples from different groups and using pre-trained visual representations to train learnable contexts.

\paragraph{Zero-shot annotation of spurious features.}
CLIP consists of a visual encoder and a text encoder, encoding image and text pairs into an aligned text space. Previous work has demonstrated CLIP's zero-shot capability: given \(K\) class names, typical text prompts can be generated, in the form of "a photo of [class]". The visual embedding of each image sample can be calculated for similarity with the text embedding of each class prompt, and it is classified into the most similar class. Directly, in order to eliminate bias towards specific spurious attributes, corresponding prompts can be generated given the class names of spurious attributes, such as "a photo of a bird [in a forest/on water]", to directly infer the attribute labels of samples.

We simply evaluated CLIP's data annotation ability for classic spurious attributes such as specific backgrounds, genders, and environments. We compared it with previous unsupervised inference methods, which can mainly be classified into two types of annotation methods:\textbf{ 1. K-means clustering} of features\cite{zhang2022contrastive,seo2022unsupervised,sohoni2020no}, where \(K\) is equal to the number of biased attributes, and UMAP feature reduction is used to improve the effect;  \textbf{2. ERM confidence }\cite{nam2020learning,liu2021just,zhang2022correct}, where the correct/misclassified samples within each category are selected as majority/minority attributes. Table \ref{tab:zero_shot} shows that the zero-shot inference of VLMs has a distinct advantage through comparison.
\begin{table*}[htbp]
\caption{Comparison of spurious attribute annotation methods. The table shows the worst-group accuracy of predicting spurious attributes across all groups, indicating that zero-shot CLIP can predict spurious attributes more accurately.}
\centering
\begin{tabular}{c|ccc|ccc}
\toprule
\multirow{2}{*}{WG(\%)} & \multicolumn{3}{c|}{CLIP ViT-L/14} & \multicolumn{3}{c}{CLIP RN50} \\
 & Waterbrids & CelebA & MetaShift & Waterbrids & CelebA & MetaShift \\
 \midrule
Kmeans & 17.9 & 62.4 & 13.3 & 57.1 & 95.5 & 16.8 \\
ERM Confidence & 16.1 & 16.5 & 40.3 & 5.4 & 12.9 & 37.2 \\
\rowcolor{gray!20}
ZS CLIP & 72.0 & 97.1 & 85.3 & 68.1 & 95.8 & 84.9 \\
\bottomrule
\end{tabular}
\label{tab:zero_shot}
\end{table*}

\paragraph{Robustness prompt tuning}
After obtaining the pseudo-labels of the sample spurious attributes, we designed the following simple reweighting algorithm. We consider the scale of samples in each group \(N_g\) and especially focus on the difficult groups, where the samples are predict poorly. Thus, we consider the average misclassify probability \(1-\overline{p}_{y|g}\) within this group to construct this weights, as follows:
\begin{equation}
\hat{w}_g \propto \frac{1}{N_g} \exp{[\eta (1-\overline{p}_{y|g})]},
\end{equation}

Here, \(\overline{p}_{y|g} = 1/N_g\sum_{(x,y)|g}\hat{p}_y\) and \(\eta > 0\) is a hyperparameter. The larger the value of \(\eta\), the more attention is paid to difficult groups; when $\eta=0$, it is a simple group-balanced re-weighting method.

To prevent the class imbalance, we further modify the weight of each group of samples by dividing it by the sum of the weights of all groups of the same class:
\begin{equation}
\hat{w}_g \leftarrow \frac{\hat{w}_g}{\sum_{k\in Y_c}\hat{w}_k\cdot N_{k}}
\end{equation}

In order to achieve stable training, the weights of each group of samples are re-estimated within each batch, and updated using exponential moving average:
\begin{equation}
w_k\leftarrow (1 - m)w_k + m\cdot \hat{w}_k
\end{equation}
Here, \(m\) is a hyperparameter and is fixed at 0.3. The final loss can automatically up-weight the difficult groups:
\begin{equation}
L(\theta)=\sum_{(x,y,g)\in D}w_g l(x,y|\theta)
\end{equation}

\section{Experiments}
\label{sec:experiment}
\subsection{Datasets, Models and Implementation}
\paragraph{Datasets} We describe the evaluated datasets:
\begin{itemize}
    \item \textbf{Waterbirds} \cite{sagawa2019distributionally} is a binary classification image dataset, with the target of classifying images into land birds (such as woodpeckers) and water birds (such as seagulls). Following the setting of \cite{sagawa2019distributionally}, the background in the images is spuriously correlated with the categories, and most land birds appear on land background (such as bamboo, forest), while most water birds appear on water background (such as the sea).
    \item \textbf{CelebA} \cite{liu2015deep} is a binary classification face dataset, where each image contains annotations of more than 40 attributes (such as hair color, eyebrows, gender, etc.). The task is to classify whether an image has blonde hair, and gender is spuriously correlated with hair color. In images of blonde hair, most of the images are female. 
    \item \textbf{MetaShift} \cite{you2024calibrating} is a cat and dog classification dataset, where images are identified as belonging to two animal categories. Following the setting of \cite{liang2022metashift}, there is a spurious correlation between the target and the background, with cats mostly appearing indoors and dogs mostly appearing outdoors.
\end{itemize}
\paragraph{Models}
Our work uses CLIP as the vision-language model and selects two popular architectures: ResNet-50 and ViT-L/14. The data augmentation for all experiments is fixed as central cropping and resizing to 224x224, without any other data augmentation, the features of all images are pre-extracted, and the vision and text encoders are frozen.
\paragraph{Baseline Methods} As baseline methods, we compare with the fine-tuning methods specific to VLMs,  CoOp \cite{zhou2022learning}, Con-Adapter \cite{zhang2022contrastive}, Orth-Proj\cite{chuang2023debiasing}, Orth-Cali\cite{chuang2023debiasing} and DPS+RNS\cite{you2024calibrating}. Meanwhile, we adapt the unimodal visual fine-tuning methods BPA\cite{seo2022unsupervised} and GDRO\cite{sagawa2019distributionally} to the prompt tuning, and AFR\cite{qiu2023simple} to the linear probe for comparison.

\paragraph{Hyperparameters}
We use the SGD optimizer and set the initial learning rate to 1e-3 and decay it cosine to 1e-4. The batch sizes for Waterbrids, CelebA, and MetaShift are 64, 1024, and 64 respectively. All tasks are trained for 50 epochs. We apply the same learning rate to all baseline methods and train for the same number of epochs, except for Con-Adapter. Additionally, we set the number of learnable contexts of prompt tuning to 6 on the Waterbirds and MetaShift datasets and 12 on the CelebA. 

\subsection{Results}
\paragraph{Improve Group Robustness.}
Table \ref{tab:main_result} shows the comparison between our method and existing methods on three datasets. First, we observe that prompt tuning CoOp and linear probe may undermine the group robustness of VLMs. Furthermore, the experimental results indicate that  DPT consistently improves the group robustness of prompt tuning, achieving the best worst-group accuracy and the smallest robustness gap compared with methods that do not rely on spurious attribute annotation. In particular, DPT significantly outperforms the best baseline methods that rely on spurious attribute annotation (such as AFR, BPA, DPS-RNS), and achieves competitive performance compared with GDRO and Con-Adapter.

\begin{table*}[htbp]
\caption{Experimental results of improving the group robustness of CLIP. The best worst-group accuracy and robustness gap \textit{without any group annotations} are \textbf{bolded}. We present the experimental results averaged over three runs, using three random seeds.}
\resizebox{\linewidth}{!}{
\begin{tabular}{c|c|ccc|ccc|ccc|ccc|ccc|ccc}
\toprule
 \multirow{3}{*}{\textbf{Dataset}}& \multirow{3}{*}{\shortstack{ Anno. \\Train/Val}} & \multicolumn{9}{c|}{\textbf{CLIP ViT-L/14}} & \multicolumn{9}{c}{\textbf{CLIP   ResNet-50}} \\
 & & \multicolumn{3}{c}{\textbf{Waterbird}} & \multicolumn{3}{c}{\textbf{CelebA}} & \multicolumn{3}{c|}{\textbf{MetaShift}} & \multicolumn{3}{c}{\textbf{Waterbird}} & \multicolumn{3}{c}{\textbf{CelebA}} & \multicolumn{3}{c}{\textbf{MetaShift}} \\
 & & WG (↑) & Avg (↑) & Gap (↓) & WG (↑) & Avg (↑) & Gap (↓) & WG (↑) & Avg (↑) & Gap (↓) & WG (↑) & Avg (↑) & Gap (↓) & WG (↑) & Avg (↑) & Gap (↓) & WG (↑) & Avg (↑) & Gap (↓) \\
\midrule
AFR & \(\times\)/\checkmark & 76.0 & 91.4 & 15.4 & 81.1 & 94.5 & 13.4 & 87.7 & 94.9 & 7.2 & 62.1 & 86.8 & 24.7 & 83.2 & 95.2 & 12.0 & 81.7 & 92.6 & 10.9 \\
BPA & \(\times\)/\checkmark & 67.7 & 83.5 & 15.7 & 86.5 & 91.6 & 5.1 & 89.4 & 96.0 & 6.7 & 64.0 & 81.0 & 17.0 & 82.6 & 92.8 & 10.2 & 85.0 & 92.1 & 7.1  \\
Con-Adapter & \(\times\)/\checkmark & 86.9 & 96.2 & 9.3 & 84.6 & 90.4 & 5.8 & 90.8 & 94.5 & 3.7 & 83.7 & 89.4 & 5.7 & 90.0 & 90.7 & 0.7 & 74.3 & 88.6 & 14.2 \\
DPS+RNS & \(\times\)/\checkmark & 88.2 & 96.8 & 8.6 & 84.8 & 87.8 & 3.0 & 93.7 & 95.5 & 1.8 & 75.4 & 84.3 & 8.9 & 81.5 & 85.0 & 3.5 & 84.6 & 90.3 & 5.7 \\ 
GDRO & \checkmark/\checkmark & 87.4 & 91.9 & 4.5 & 88.9 & 91.0 & 2.1 & 92.8 & 96.3 & 3.5 & 83.7 & 88.2 & 4.5 & 88.5 & 92.4 & 3.8 & 86.8 & 93.8 & 6.9\\
\midrule
Zero-shot & \(\times\)/\(\times\) & 43.2 & 84.0 & 40.8 & 82.9 & 85.8 & 2.9 & 90.8 & 96.2 & 5.5 & 40.0 & 77.0 & 37.0 & 62.7 & 73.5 & 10.9 & 84.6 & 95.1 & 10.5 \\
Linear Probe & \(\times\)/\(\times\) & 65.9 & 97.6 & 31.7 & 28.3 & 94.7 & 66.4 & 87.9 & 95.9 & 8.0 & 7.9 & 93.5 & 85.6 & 11.9 & 94.7 & 82.8 & 23.1 & 71.6 & 48.6 \\
CoOp & \(\times\)/\(\times\) & 58.4 & 90.8 & 32.4 & 23.9 & 94.4 & 70.6 & 90.8 & 96.6 & 5.8 & 49.1 & 84.3 & 35.3 & 32.2 & 95.3 & 63.1 & 83.1 & 94.1 & 11.0 \\
Orth-Proj & \(\times\)/\(\times\) & 61.4 & 86.4 & 25 & 71.1 & 87.6 & 16.5 & 92.1 & 94.6 & 2.5 & 48.1 & 83.6 & 35.5 & 61.4 & 86.4 & 25 & 84.6 & 90.3 & 5.7 \\
Orth-Cali & \(\times\)/\(\times\) & 68.8 & 84.5 & 15.7 & 76.1 & 87 & 10.9 & 91.6 & 95.5 & 3.9 & 74 & 78.7 & 4.7 & 82.2 & 84.4 & 2.2 & 81.5 & 89.5 & 8.0 \\
\rowcolor{gray!20}
\textbf{DPT} & \(\times\)/\(\times\) & \textbf{86.5} & 90.2 & \textbf{3.7} & \textbf{88.4} & 89.7 & \textbf{1.3} & \textbf{93.8} & 96.0 & \textbf{2.1} & \textbf{81.2} & 83.3 & \textbf{2.2} & \textbf{90.0} & 91.0 & \textbf{1.0} & \textbf{87.7} & 93.4 & \textbf{5.7} \\
\bottomrule
\end{tabular}
}
\label{tab:main_result}
\end{table*}

\begin{figure*}[!htbp]
\centering
\subfloat[Waterbirds]{
        \label{fig:grad_cam_waterbirds}
        \includegraphics[width=0.48\linewidth]{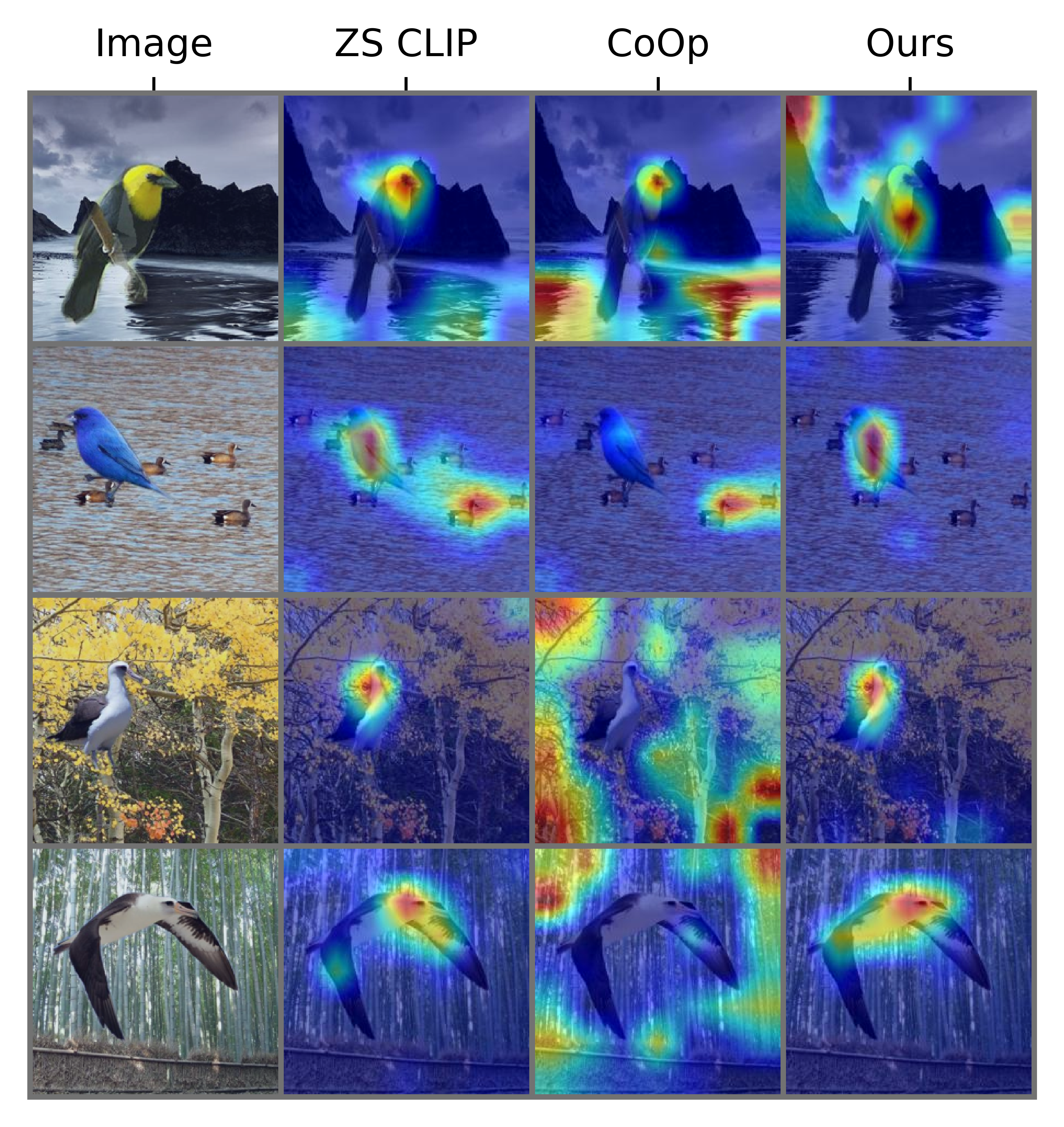}
        }\hfill
\subfloat[CelebA]{
        \label{fig:grad_cam_celeba}
        \includegraphics[width=0.48\linewidth]{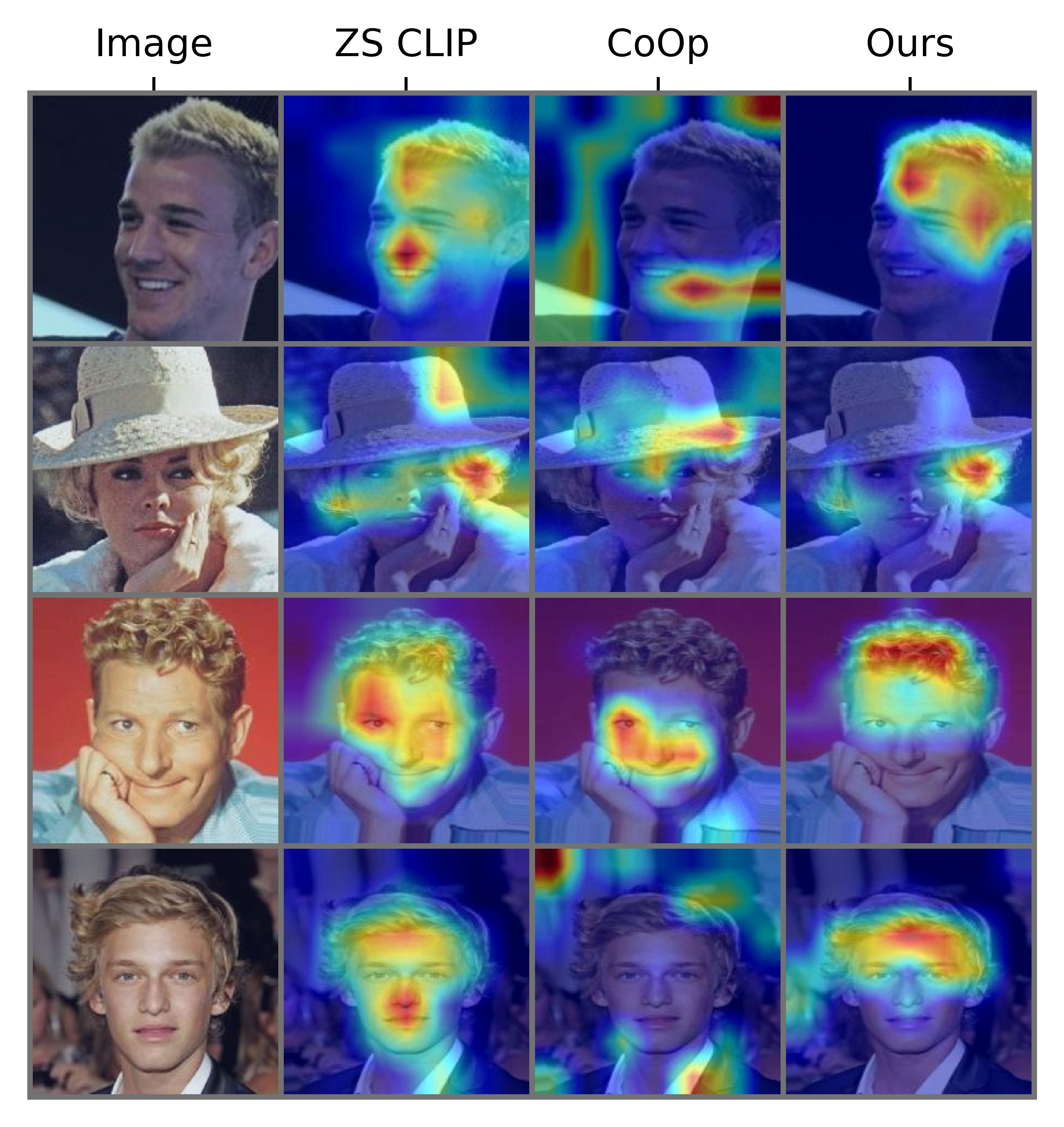}
        }
\caption{We use Grad-CAM \cite{selvaraju2017grad} to visualize attention for different approaches (Zero-shot CLIP, CoOp, and Ours) using CLIP-ResNet-50 for Waterbirds and CelebA datasets.}
\label{fig:grad_cam}
\end{figure*}

In addition, we also visualize different method's attention map using Grad-CAM\cite{selvaraju2017grad} in CLIP-ResNet-50. As shown in Fig. \ref{fig:grad_cam}, the pretrained CLIP can focus on the birds but relies on facial features other than hair color; CoOp does not focus on the targets (birds and hair color) and depends on other features. However, our method pays more attention to utilize the target features (birds and blonde-hair).

\paragraph{Transfer across VLMs}
To further evaluate the effectiveness of our method, we evaluate the improvement in robustness across different foundation models. In the Table \ref{tab:model}, we run experiments on two typical VLMs, ALIGN\cite{jia2021scaling} and BLIP\cite{li2022blip}, trained with multimodal contrastive losses, as well as on additional model architectures RN101 and ViT-B/32 of CLIP\cite{radford2021learning}. We find that DPT consistently improves the group robustness of prompt tuning and achieves comparable worst-group accuracy and robustness gap to the current state-of-the-art robustness method. 
\begin{table*}[!htbp]
    \caption{On the three datasets, DPT consistently improves the group robustness of prompt tuning on BLIP, ALIGN, and two additional variants of CLIP. In addition, when compared with the state-of-the-art GDRO that requires group annotations, DPT achieves competitive performances.
    }
    \centering
    \resizebox{\linewidth}{!}{
    \begin{tabular}{c|l|ccc|ccc|ccc}
    \toprule
        \multirow{2}{*}{\textbf{Backbone}} & \multirow{2}{*}{\textbf{Method}} & \multicolumn{3}{c|}{\textbf{Waterbird}} & \multicolumn{3}{c|}{\textbf{CelebA}} & \multicolumn{3}{c}{\textbf{Metashift}} \\
         &  & WG (↑) & Avg (↑) & Gap (↓) & WG (↑) & Avg (↑) & Gap (↓) & WG (↑) & Avg (↑) & Gap (↓)  \\ 
         \midrule
        \multirow{4}{*}{BLIP} & Zero-shot & 15.1 & 60.0 & 44.9 & 64.5 & 74.4 & 9.9 & 90.1 & 94.4 & 4.3 \\
         & CoOp & 45.5 & 81.8 & 36.4 & 28.9 & 95.1 & 66.2 & 87.7 & 95.7 & 8.0 \\
         & GDRO & {\underline{80.1}} & 87.0 & {\underline{6.9}} & \textbf{91.6} & 91.9 & \textbf{0.3} & \underline{90.8} & 95.2 & \underline{4.4} \\
         \rowcolor{gray!20}
         & DPT(ours) & \textbf{83.1} & 88.5 & \textbf{5.4} & {\underline{90.8}} & 91.5 & {\underline{0.6}} & {\textbf{90.8}} & 95.2 & {\textbf{4.4}} \\
          
        \midrule
        \multirow{4}{*}{ALIGN} & Zero-shot & 53.7 & 68.0 & 14.3 & 75.9 & 80.5 & 4.6 & \textbf{89.2} & 92.7 & \textbf{3.4} \\
         & CoOp & 26.3 & 82.2 & 55.9 & 22.2 & 94.1 & 71.9 & 81.5 & 92.9 & 11.4 \\
         & GDRO & {\underline{66.5}} & 76.4 & {\underline{9.9}} & \textbf{88.9} & 89.8 & \textbf{0.9} & 66.5 & 76.4 & 9.9  \\
         \rowcolor{gray!20}
         & DPT(ours) & \textbf{76.7} & 84.8 & \textbf{8.1} & {\underline{87.8}} & 88.8 & {\underline{1.1}} & 84.6 & 94.3 & 9.7 \\

        \midrule
        \multirow{4}{*}{CLIP   RN101} & Zero-shot & 45.9 & 70.8 & 24.9 & 69.6 & 78.9 & 9.3 & 89.0 & 94.4 & 5.4 \\
         & CoOp & 54.7 & 84.7 & 30.0 & 28.9 & 95.2 & 66.3 & 83.1 & 94.4 & 11.3 \\
         & GDRO & \textbf{84.4} & 89.2 & \textbf{4.8} & {\underline{90.6}} & 91.8 & {\underline{1.2}} & {\underline{89.2}} & 93.2 & {\underline{4.0}} \\
         \rowcolor{gray!20}
         & DPT(ours) & {\underline{82.4}} & 87.6 & {\underline{5.2}} & \textbf{90.7} & 91.2 & \textbf{0.5} & \textbf{90.8} & 93.1 & \textbf{2.4} \\

        \midrule
        \multirow{4}{*}{CLIP   ViT-B/32} & Zero-Shot & 45.5 & 69.7 & 24.2 & 78.1 & 81.5 & 3.4 & 84.6 & 94.1 & 9.4 \\
         & CoOp & 40.3 & 81.3 & 41.0 & 32.2 & 95.2 & 63.0 & 81.5 & 93.6 & 12.1 \\
         & GDRO & {\underline{80.8}} & 86.4 & {\underline{5.6}} & \textbf{91.1} & 91.7 & {\underline{0.6}} & {\underline{84.6}} & 93.7 & {\underline{9.1}}\\
         \rowcolor{gray!20}
         & DPT(ours) & \textbf{81.6} & 86.5 & \textbf{4.9} & {\underline{91.0}} & 91.4 & \textbf{0.3} & \textbf{87.7} & 93.4 & \textbf{5.7} \\
        \bottomrule
    \end{tabular}
    }
    \label{tab:model}
\end{table*}

\paragraph{Robustness to Unimodal Vision Model}
In the previous section, we introduced how DPT significantly improves the group robustness of the foundation models. We now  evaluate the efficiency of DPT. For the evaluation, we compared it with the current robustness methods on two popular datasets, Waterbirds and CelebA. These methods train the ImageNet-pretrained ResNet50 and require group annotations in the training set or validation set. The Table \ref{tab:resnet50} shows that DPT achieve comparable performance to current robustness methods while only training less than 1\% of the parameters. In particular, DPT outperforms all state-of-the-art methods on CelebA. These indicate that DPT can not only improve the group robustness of VLMs but also achieve competitive results with robust vision models through efficient fine-tuning.

\begin{table*}[htbp]
    \caption{On the popular datasets Waterbirds and CelebA, DPT achieves near state-of-the-art worst-group accuracy and robustness gap with trained $<$1\% paramters. We report the numbers from original papers, and the first/second best performance \textbf{bolded}/\underline{underlined}.}
    \centering
    \resizebox{\linewidth}{!}{
    \begin{tabular}{l|c|c|c|c|ccc|ccc}
    \toprule
        \multirow{2}{*}{Model} & \multirow{2}{*}{\shortstack{Group Info\\ Train/Val}} & \multirow{2}{*}{\# Trained Params} & \multirow{2}{*}{\%  Params}&\multirow{2}{*}{Method} & \multicolumn{3}{c|}{Watebrids} & \multicolumn{3}{c}{CelebA} \\ 
         &  &  & & & WG (↑) & Avg (↑) & Gap (↓) & WG (↑) & Avg (↑) & Gap (↓)  \\ 
         \midrule
        \multirow{8}{*}{Resnet50} & \checkmark/\checkmark & 25557032 & 100 & LISA\cite{yao2022improving} & 89.2 & 91.8 & 2.6 & 89.3 & 92.4 &3.1\\ 
         & \checkmark/\checkmark & ~ & ~ & GDRO\cite{sagawa2019distributionally} & 91.4 & 93.5 & 2.1 & 88.9 & 92.9 & 4.0\\ 
         & \checkmark/\checkmark & ~ & ~ & DFR\cite{kirichenko2022last} & \textbf{92.9} & 94.2 & \textbf{1.3} & 88.3 & 91.3 &3.0\\ 
         & \(\times\)/\checkmark & ~ & ~ & ERM & 72.6 & 97.3 & 24.7 & 47.2 & 95.6 & 48.4\\ 
         & \(\times\)/\checkmark & ~ & ~ & EIIL\cite{creager2021environment} & 78.6 & 96.9 & 18.3 & 81.7 & 85.7 & 4.0\\ 
         & \(\times\)/\checkmark & ~ & ~ & JTT\cite{liu2021just} & 86.7 & 93.3 & 6.6 & 81.1 & 88.0 &6.9\\ 
         & \(\times\)/\checkmark & ~ & ~ & CNC\cite{zhang2022correct} & 88.5 & 90.9 & 2.4 & 88.8 & 89.9 &1.1\\ 
         & \(\times\)/\checkmark & ~ & ~ & AFR\cite{qiu2023simple} & 90.4 & 94.2 & 3.8 & 82.2 & 91.3 &9.1\\ 
        \midrule
        \rowcolor{gray!20}
        CLIP ResNet-50 & \(\times\)/\(\times\) & $<$4096 & $<$0.2 & DPT(ours) & 81.2 & 83.3 & 2.2 & \textbf{90.0} & 91.0 &\textbf{1.0} \\ 
        \rowcolor{gray!20}
        CLIP ViT-L/14 & \(\times\)/\(\times\) & ~ & ~ & ~ & 86.5 & 90.2 & 3.7 & 88.4 & 89.7 &1.3\\ 
    \bottomrule
    \end{tabular}
    }

    \label{tab:resnet50}
\end{table*}

\section{Analyses}
\label{sec:ablation}
\begin{figure*}[htbp]
  \subfloat[CelebA]{
        \label{fig:eta_celeba}
        \includegraphics[width=0.3\linewidth]{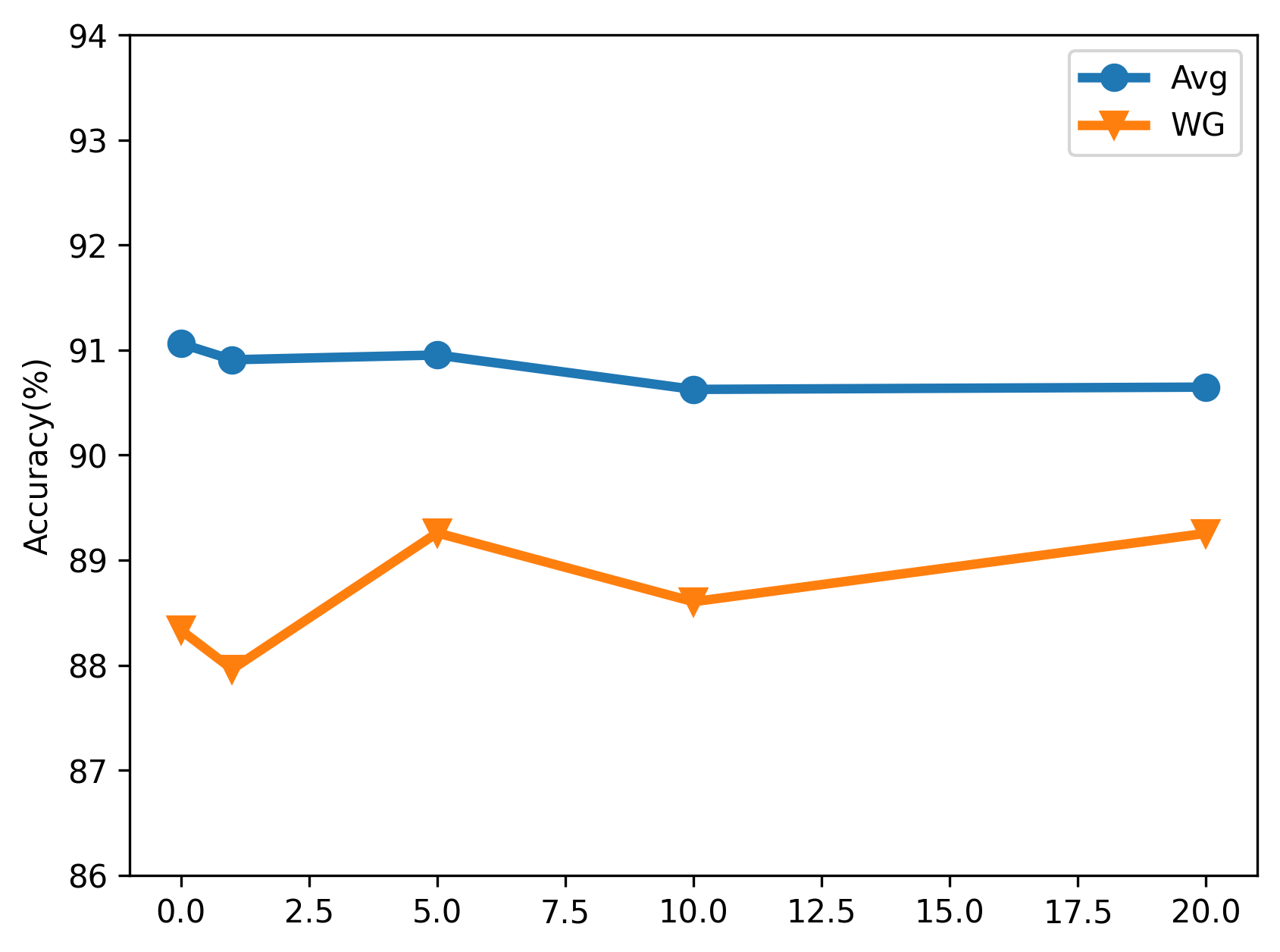}
        }\hfill
  \subfloat[Waterbirds]{
        \label{fig:eta_waterbirds}
        \includegraphics[width=0.3\linewidth]{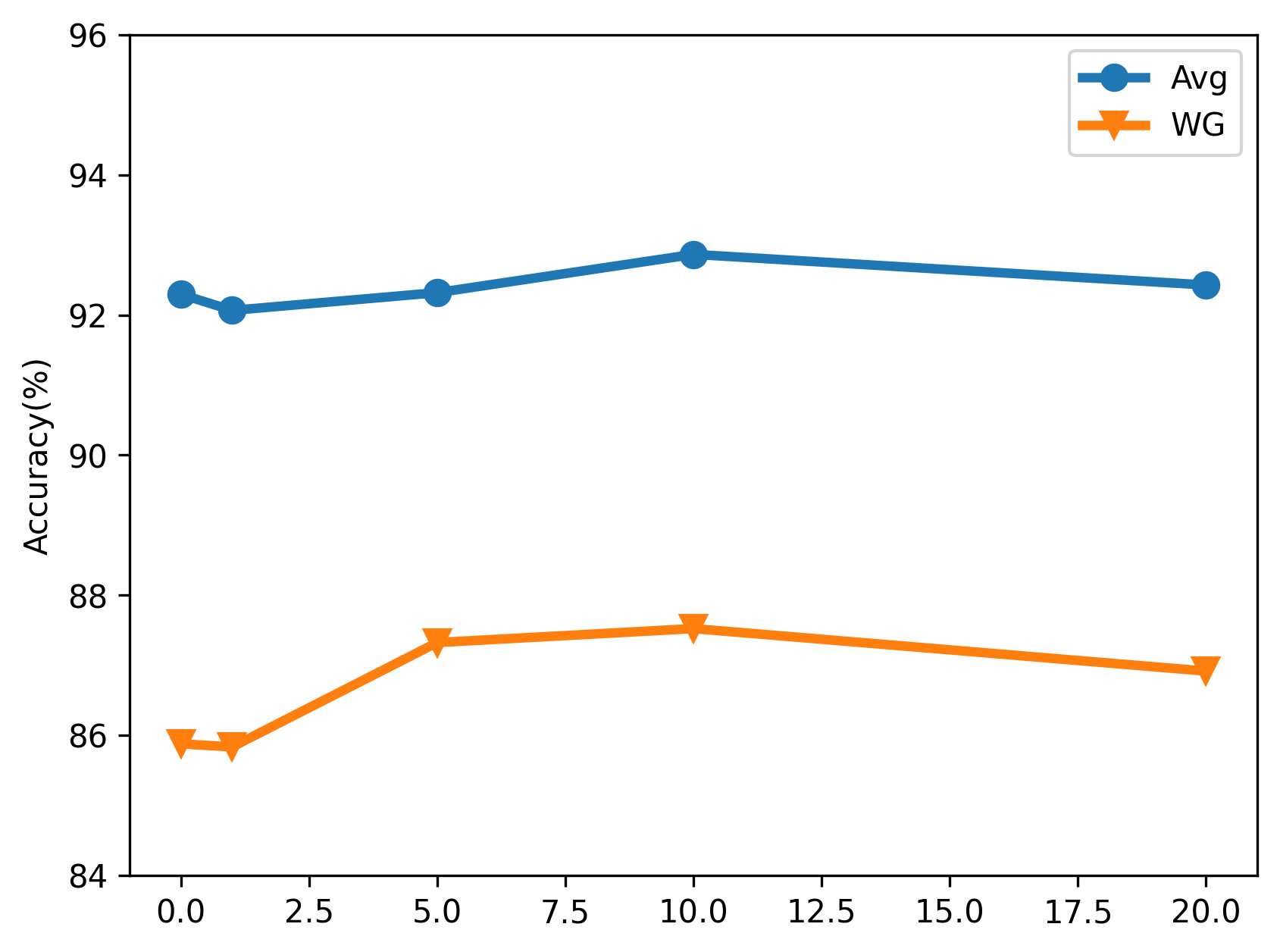}
        }\hfill
  \subfloat[MetaShift]{
        \label{fig:eta_metashift}
        \includegraphics[width=0.3\linewidth]{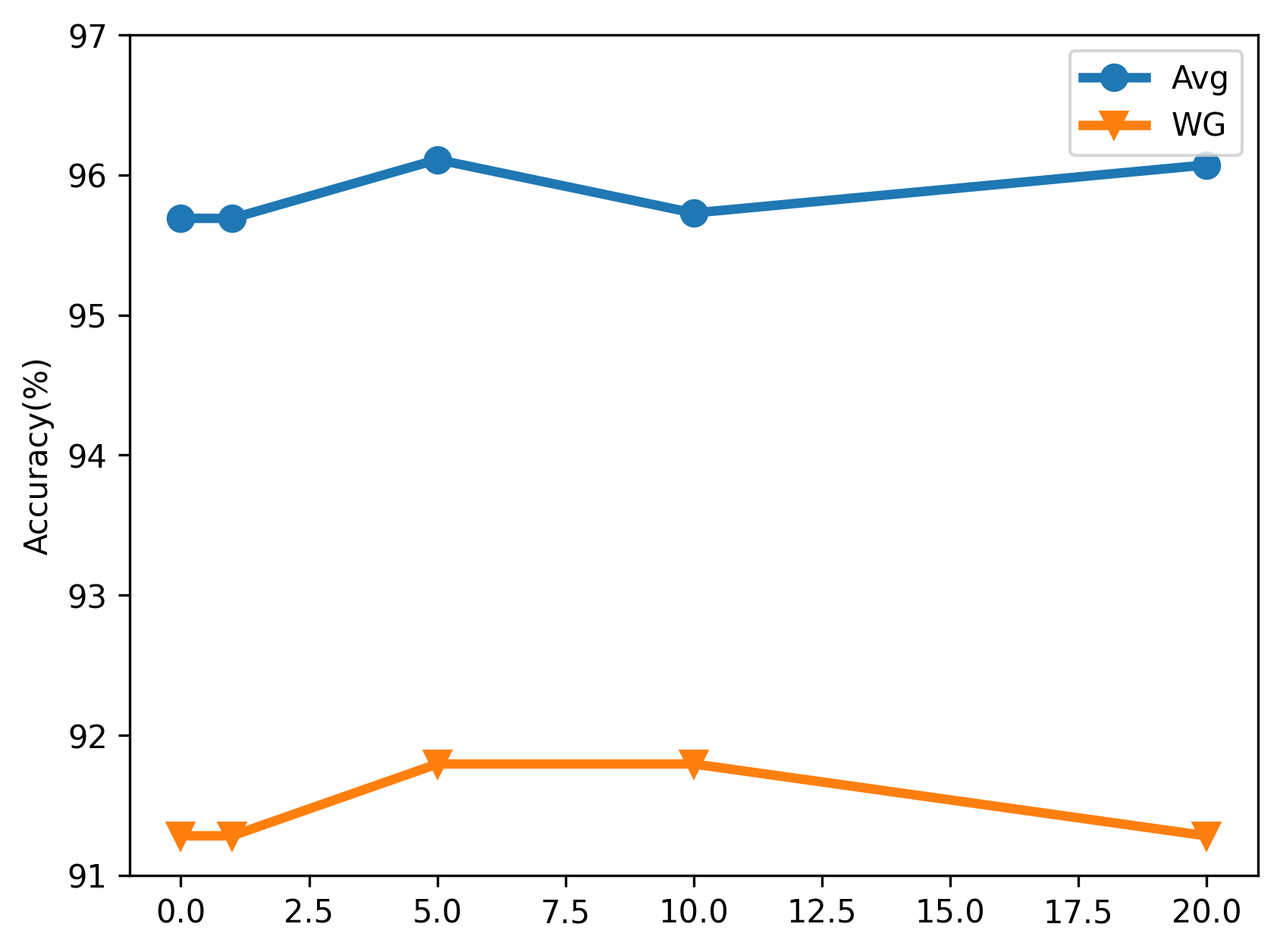}
        }
  \caption{Analysis on the sensitivity of hyper-parameters $\eta$ on CLIP ViT-L/14. We run experiments and report means over three seeds.}
  \label{fig:eta}
\end{figure*}

\paragraph{Effect of \(\eta\)}
We investigate the robustness of DPT to the selection of the hyperparameter \(\eta\) on three datasets. As mentioned before, when \(\eta > 0\), the training difficulty of each group is dynamically considered, and we do not expect DPT to be sensitive to the hyperparameter. The ablation results shown in the Fig. \ref{fig:eta} indicate that \(\eta\in[5, 10]\) can achieve good performance, and an extremely large \(\eta\) may have a negative effect. We use three random seeds and get the average of the results.

\paragraph{Effect of Zero-Shot Annotation.}
In the previous section, we briefly evaluated the bias annotation ability of CLIP. We now compare different bias annotations in terms of improving group robustness. As shown in Table \ref{tab:anno}, ZS CLIP can achieve the best group robustness and obtain results close to ground-truth annotations; and K-means clustering also leads to a significant improvement.
\begin{table}[htbp]
    \caption{We evaluate the worst-group accuracy of using different spurious attribute annotation methods. We conduct experiments on CLIP ViT-L/14, using the same re-weighting algorithm and hyperparameter settings, and only changing the group annotations of the training/validation sets.}
    \centering
    \begin{tabular}{c|ccc|c}
    \toprule
        WG(\%) & Waterbrids & CelebA & MetaShift & mean  \\ 
    \midrule
        Kmeans & 47.8 & 83.9 & 92.3 & 74.7  \\
        ERM Confidence & 63.2 & 55.6 & 91.6 & 70.1  \\ 
    \rowcolor{gray!20}
        ZS CLIP & 86.5 & \textbf{88.4} & \textbf{93.8} & \textbf{89.6}  \\ 
    \midrule
        True Label & \textbf{87.1 }& 88.3 & 92.3 & 89.2  \\
    \bottomrule
    \end{tabular}
    \label{tab:anno}
\end{table}

\paragraph{Computationally Efficient.}
We compare the time consumption of our method with that of the baseline methods, in Table \ref{tab:time}. For a fair comparison, we pre-extracted and fixedly stored the visual features of all datasets after they were encoded by the model. We conducted experiments on three datasets, and the experimental results show that the time consumption of our method is acceptable.
\begin{table}[htbp]
    \caption{Training time comparison on CLIP ViT-L/14}
    \centering
    \begin{tabular}{l|ccc}
    \toprule
        Time/min & Waterbrids & CelebA & MetaShift  \\ 
    \midrule
        GDRO & 5.2  & 10.5  & 3.6   \\
        AFR & 3.3  & 11.7  & 4.2   \\ 
        BPA & 7.0  & 15.8  & 3.8   \\
        Con-Adapter & 30.0  & 270.0  & 20.0   \\ 
    \rowcolor{gray!20}
        Ours & 4.7  & 9.3  & 3.5   \\
    \bottomrule
    \end{tabular}
    \label{tab:time}
\end{table}

\section{Conclusion}
\label{sec:conc}
In this work, we focus on the robustness of prompt tuning in Vision-Language Models (VLMs) when dealing with spurious correlations. To address this issue, we propose a simple yet effective robust fine-tuning method. It first infers spurious attributes using text descriptions and the model's zero-shot capabilities, and eliminates the model's dependence on spurious correlations through dynamic re-weighting. Experimental results show that we have significantly improved group robustness. 

\section{Limitation and Discussion}
Our work has several future directions. Firstly, our method is targeted at specific spurious correlations and requires knowledge of the names of spurious attributes. Therefore, the main applicable scenarios may be those addressing fairness issues (such as social biases in age, gender, and skin color) or problems with expert prior knowledge. Automatically discovering spurious correlations in models is a direction for improvement. Secondly, our method can only eliminate spurious attributes that are recognizable by VLMs and can be appropriately described by text. It cannot eliminate spurious correlations that are difficult to identify. This limits its ability to address unrecognizable robustness issues such as adversarial perturbations. However, our method is computationally efficient, exhibits excellent performance, and can be easily extended from a single spurious attribute to multiple spurious attributes. We believe it can meet the needs of more complex real-world applications.

\bibliographystyle{unsrt}
\bibliography{neurips_2024}



\end{document}